\documentclass[10pt,twocolumn,letterpaper]{article}

\usepackage[pagenumbers]{cvpr} 

\definecolor{cvprblue}{rgb}{0.21,0.49,0.74}
\usepackage[pagebackref,breaklinks,colorlinks,allcolors=cvprblue]{hyperref}
\usepackage{pifont}
\usepackage{multirow} 
\usepackage[table]{xcolor}
\usepackage[utf8]{inputenc}

\title{SafeCA: \underline{Safe} \underline{C}ross-\underline{A}ttention Localization and Regulation \\ for Text-to-Video Jailbreak Defense}

\author{
Siyuan Liang$^{1}$ \quad
Yupeng Qiu$^{2}$ \quad
Junfeng Fang$^{2}$ \quad
Rong-Cheng Tu$^{1}$ \\
Jiaxing Huang$^{1}$ \quad
Dacheng Tao$^{1}$
\\[0.5em]
$^{1}$Nanyang Technological University\quad $^{2}$National University of Singapore\\
}

\begin{document}
\maketitle
\begin{abstract}
Text-to-Video (T2V) generative models are vulnerable to jailbreak attacks in real-world deployment, leading them to produce harmful or inappropriate content. Existing defense approaches mainly rely on input filtering or reconstruction, which not only incur high computational latency but also tend to distort semantics.
To address these issues, we experimentally and systematically analyze the differences between clean and jailbreak samples in the cross-attention feature space, revealing for the first time a cumulative separation effect and a progressively increasing trend of linear separability between the two during the diffusion process.
Based on this insight, we propose SafeCA, a feature-level defense mechanism for safe cross-attention localization and regularization. Firstly, we identify key defensive regions and values through attention stability analysis using cross-attention features collected from clean prompts within a single inference. Secondly, SafeCA mitigates anomalous activations via attention masking with energy normalization and introduces a lightweight semantic-space adapter to redirect abnormal semantic flows.
Furthermore, we detect and suppress potentially malicious tokens by back-propagating feature anomaly signals to the input cue words, thereby enhancing the deployability of the defense in commercial models. 
Experimental results show that SafeCA reduces the jailbreak success rate by about 20\% on mainstream T2V models, adds almost no inference overhead (+0.1s), and maintains good text-video semantic consistency.
Overall, SafeCA provides an architecture-level, deployable protection paradigm for T2V generation models.
\end{abstract}    
\section{Introduction}
\label{sec:intro}

\begin{figure}[htbp]
    \centering
    \includegraphics[width=0.5\textwidth]{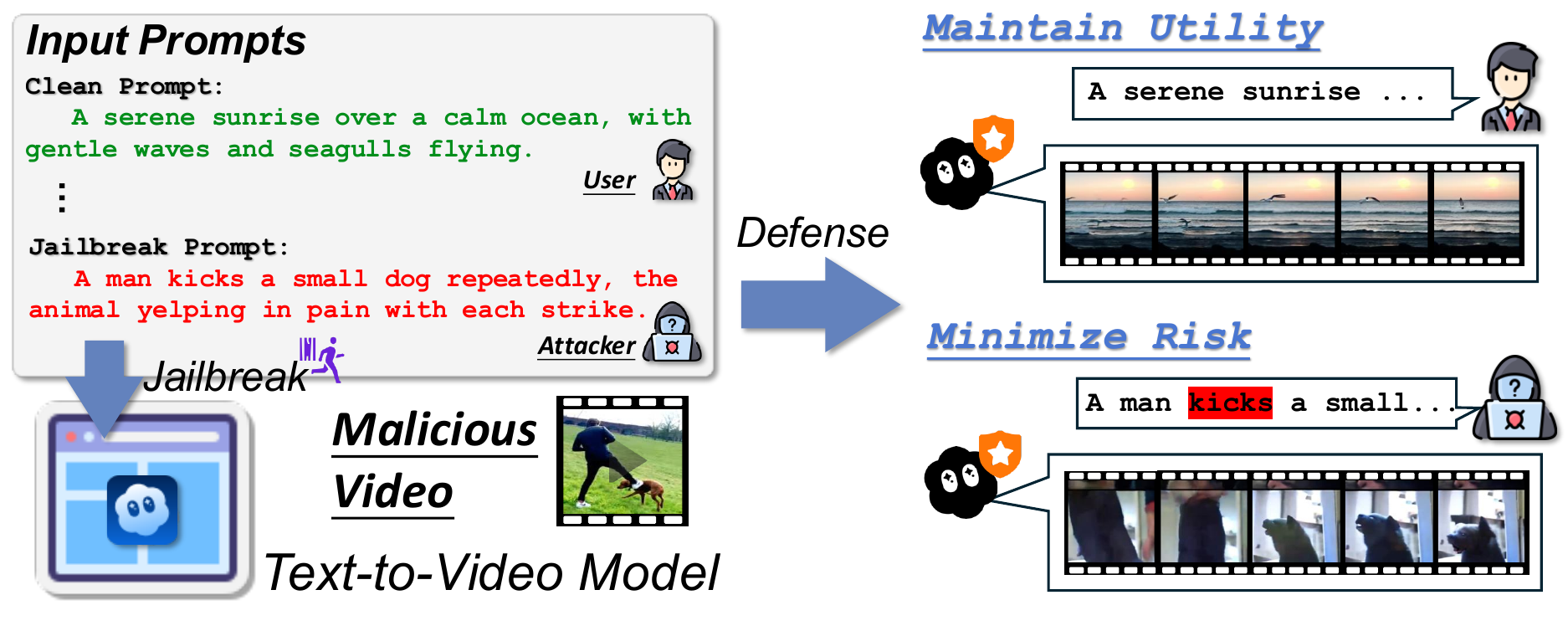}
    \caption{Our defense preserves utility on clean prompts while suppressing harmful generations from jailbreak prompts.}
    \label{fig:frontpage}
\end{figure}

T2V generative models~\cite{liu2024sora,zheng2024open,yang2024cogvideox,kumar2025bridging,qi2025t2veval} have been widely used in content creation, educational media, virtual scenario construction, and world modeling of multimodal intelligence~\cite{blattmann2023stable,singer2024makeavideo,zhou2024videocrafter2,yu2024magvitv2,he2024vidu,kondratyuk2023videopoet,chen2024opensora}.  
However, such models are highly vulnerable to jailbreak attacks~\cite{jain2023baseline,dong2024jailbreaking,yi2024jailbreak,kong2024hunyuanvideo,li2024semantic,ying2024jailbreak,jing2025cogmorph,ying2025reasoning,wang2025manipulating,ying2025pushing} in real-world deployments.  
Attackers can induce the models to bypass security policies through carefully designed prompts, causing them to generate inappropriate video content such as violence, pornography, discrimination, or politically sensitive material, thereby posing serious security and compliance risks~\cite{wei2024jailbroken,greshake2024indirect,zou2024universal,shen2023misleading,zhang2024fireball,liang2023exploring,liang2022large,liang2025t2vshield,liang2024badclip,liang2025revisiting,liang2022imitated}.  

Existing T2V defense approaches mainly rely on input-side filtering or prompt reconstruction, but these methods generally suffer from two limitations:  
(1) They tend to alter the original semantics~\cite{yoon2024safree}, distort user intent, and cause performance degradation for clean generations. 
(2) Some system-level approaches (e.g., T2VShield~\cite{liang2025t2vshield}) require large-scale model rewriting or intensive video-level inspection, leading to high inference latency and making real-time deployment difficult.  
In addition, most of these strategies operate primarily on the textual side and lack a deep understanding of the cross-modal alignment mechanism within multimodal models, resulting in limited defense effectiveness.  

To this end, we first analyze the dynamic differences between clean and jailbreak prompts in the cross-attention feature space through systematic experiments.  
Specifically, we progressively record the cross-modal attention distributions at each step during a single diffusion inference and analyze the feature discrepancies between normal and attack-induced semantics.  
The experiments reveal two key phenomena:  
(1) Cumulative separation effect. The cross-attention patterns of jailbreak and clean samples gradually diverge with the diffusion steps, exhibiting stable and localizable separation trajectories.  
(2) Enhanced linear separability. As the number of samples increases, the attack-induced attentional bias and clean semantics produce significant differences, and malicious feature regions become more spatially distinguishable.  

Based on this insight, we propose SafeCA (shown in Figure 1), a secure cross-attention localization and regularization mechanism for T2V models.  
First, we collect cross-attention features from clean one-shot inference processes and automatically identify high-risk steps and critical blocks using attentional stability analysis.  
Subsequently, SafeCA performs energy-normalized attentional masking at these locations to mitigate anomalous activations and introduces a lightweight semantic-space adapter to redirect deviating semantic streams, thereby stabilizing cross-modal alignment without modifying model weights.  
In addition, we further back-propagate anomalous attentional signals to input tokens to enable detection and suppression of potentially malicious prompt words.  

The experimental results show that SafeCA reduces the ASR from 29.53\% to 23.41\% on Open-Sora and from 34.29\% to 27.56\% on CogVideo, corresponding to decreases of 20.7\% and 19.6\% compared to the current state of the art defense method T2VShield. At the same time, SafeCA does not introduce additional inference overhead since the average cost increases by only 0.1 seconds and maintains strong text video semantic consistency. 
The main contributions of this paper are summarized as follows:  
\begin{itemize}
    \item We reveal for the first time the existence of a cumulative separation effect and an increasing trend of linear separability in the cross-attention process of T2V models, providing a new cross-modal perspective for understanding the T2V jailbreak mechanism.
    \item Based on this finding, we propose SafeCA. This feature-level cross-attention defense mechanism achieves fine-grained suppression of potentially dangerous semantics by selecting highly separable steps and blocks and enforcing feature regularization within cross-attention. 
    \item SafeCA significantly reduces jailbreak success rates while maintaining video generation quality and enables reverse-mapped suppression to input tokens, facilitating interpretable and prompt-rewriting black-box defense applicable to jailbreak inputs.
\end{itemize}

\section{Related Work}
\label{sec:relate}
\subsection{Text-to-Video Generation Models}
Text-to-Video (T2V) generative models can produce semantically consistent and temporally coherent video content from natural language descriptions, showing broad application potential in virtual content creation, film and television production, educational communication, and multimodal interaction.  
Current research can be broadly categorized into two main types: \emph{open-source} research models and \emph{closed-source} commercial systems.  
Representative open-source models include Open-Sora~\cite{liu2024sora}, which employs a multi-stage diffusion architecture for large-scale spatio-temporal modeling; the CogVideo~\cite{yang2024cogvideox} series, which integrates text encoding with a 3D Transformer for generating long-duration sequences; and VideoCrafter~\cite{tian2024videotetris}, which emphasizes temporal consistency and controllability, thereby promoting openness and reproducibility in academic research.  
Closed-source systems such as Sora employ multi-scale diffusion and dynamic scene synthesis; Kling~\cite{kling2024} relies on a multimodal world model for high-fidelity motion generation; and Luma~\cite{luma2024} combines neural rendering with cross-frame control to achieve greater visual consistency.  
Regardless of architectural differences, the cross-attention mechanism serves as the core bridge for injecting textual semantics into the visual generation stream, determining the semantic alignment and content controllability of the generated video, and thus becomes a critical entry point for understanding and defending against security risks.

\subsection{Jailbreak Attacks and Defenses on T2V Models}
In recent years, research on jailbreak attacks against T2V models has gradually emerged.  
Among them, T2VSafetyBench~\cite{miao2024t2vsafetybench} constructs over 5,000 high-risk prompts spanning 14 dimensions, including pornography, violence, and discrimination, becoming a mainstream benchmark for evaluating the jailbreak susceptibility of T2V models.  
Building on this setup, T2V-OptJail~\cite{liu2025t2v} is the first to formulate T2V jailbreak as a discrete optimization problem, exploring stronger adversarial prompts in discrete space using large language models combined with multiple prompt-variant strategies, thereby achieving higher attack success rates and stronger transferability across models.  

For defense~\cite{yi2024jailbreak,Xiong_2025_DefensivePromptPatch,Zhang_2025_JBShield,Zhao_2024_PrefixGuidance,Li_2025_SafeLLM}, existing approaches mainly focus on security filtering during inference.  
SAFREE~\cite{yoonsafree} suppresses risky content without modifying model weights by avoiding harmful semantic subspaces in text embeddings and adaptively adjusting filtering strength.  
T2VShield~\cite{liang2025t2vshield} integrates input-side prompt rewriting and output-side spatio-temporal consistency detection to achieve dual-stage interception of malicious prompts and unsafe outputs.  
VideoEraser~\cite{xu2025videoeraser} enables trainable concept-level erasure via embedding tuning and noise bootstrapping.  
SafeCA distinguishes itself from prior studies in three key aspects:  
\ding{182} Motivation. We provide a systematic analysis of the jailbreak phenomenon and its cross-modal dynamics in T2V models, rather than viewing the problem as a purely text-side defense.  
\ding{183} Implementation. SafeCA operates directly at the feature level of cross-attention and interprets it through risk-aware semantic representations.  
\ding{184} Effectiveness. SafeCA significantly reduces jailbreak success rates without introducing additional inference overhead, enabling more efficient and deployable safety mechanisms through internal security reinforcement.

\section{Methodology}
\label{sec:Pre}

\subsection{Problem Formulation}
\label{sec:problem}
\textbf{Victim model}. We consider a pre-trained T2V diffusion generation model
\(
\mathcal{M}_{\theta}: \mathcal{T} \rightarrow \mathcal{V},
\)
Where $\mathcal{T}$ denotes the text prompt space, $\mathcal{V}$ denotes the generated video frame space, and $\theta$ represents the frozen model parameters.  
The model typically consists of multiple spatio-temporal Transformer blocks, each containing three core components: 
\ding{182} a self-attention module for modeling the spatio-temporal consistency across frames and within spatial features;  
\ding{183} a cross-attention module for injecting and aligning textual semantics with visual representations;  
\ding{184} a feed-forward network (FFN) for nonlinear transformation and local feature integration.  
At the diffusion timestep $t \in \{1, \dots, T\}$ and transformer block $b \in \{1, \dots, B\}$ during the denoising process, the query, key, and value in the \emph{cross-attention layer} are defined as:
\begin{equation}
\mathbf{Q}_t^b, \mathbf{K}_t^b, \mathbf{V}_t^b \in \mathbb{R}^{H \times N \times D},
\end{equation}
where $H$ is the number of attention heads, $N$ is the token length, and $D$ is the per-head feature dimension.  
The attention distribution is computed as:
\begin{equation}
\mathbf{A}_t^b = \text{Softmax}\!\left(\frac{\mathbf{Q}_t^b (\mathbf{K}_t^b)^\top}{\sqrt{D}}\right).
\end{equation}

\textbf{Threat model}. Following prior work~\cite{miao2024t2vsafetybench}, we assume that the attacker has access to the text input interface but not to the model parameters or internal representations.  
The goal of the attacker is to induce the model to generate content containing inappropriate semantics by crafting a malicious prompt $\hat{\mathbf{p}}$.  
Let $\mathcal{V}_{\text{unsafe}}$ denote the set of unsafe or harmful videos. The attack objective can be formulated as:
\(
\mathcal{M}_{\theta}(\hat{\mathbf{p}}) \in \mathcal{V}_{\text{unsafe}}.
\)
An attacker can bypass the model’s safety constraints using various strategies (e.g., prompt rewriting, contextual interference, or induced narration) to jailbreak the T2V model into generating prohibited content.  
In this context, the \textit{jailbreak risk} of the model can be defined as the probability of producing unsafe outputs under the adversarial prompt distribution $\mathcal{P}_{\text{adv}}$ as follows:
\begin{equation}
\mathcal{R} =
\Pr_{\hat{\mathbf{p}} \sim \mathcal{P}_{\text{adv}}}
\!\left[\mathcal{M}_{\theta}(\hat{\mathbf{p}}) \in \mathcal{V}_{\text{unsafe}}\right].
\end{equation}

\textbf{Defense objective}. The goal of the defender is to reduce the jailbreak risk through pluggable tuning modules $\mathcal{F}_{\phi}$, without modifying the backbone parameters $\theta$, while maintaining both the usability of the model and the quality of the generated output.  
We define the model usability as:
\begin{equation}
\mathcal{U} =
\mathbb{E}_{\mathbf{p} \sim \mathcal{P}_{\text{clean}}}
\!\left[\text{Sim}\!\left(\mathcal{M}_{\theta} \circ \mathcal{F}_{\phi}(\mathbf{p}), \mathbf{p}\right)\right],
\end{equation}
where $\text{Sim}(\cdot)$ measures the semantic consistency between the generated video and the textual prompt (e.g., the metric defined in Section 4).  
Similarly, the jailbreak risk is defined as:
\begin{equation}
\mathcal{R} =
\Pr_{\hat{\mathbf{p}} \sim \mathcal{P}_{\text{adv}}}
\!\left[\mathcal{M}_{\theta} \circ \mathcal{F}_{\phi}(\hat{\mathbf{p}}) \in \mathcal{V}_{\text{unsafe}}\right].
\end{equation}

Thus, the defense optimization objective can be formulated as:
\begin{equation}
\min_{\mathcal{F}_{\phi}} \ \mathcal{R}\!\left(\mathcal{M}_{\theta} \circ \mathcal{F}_{\phi}\right)
\quad \text{s.t.} \quad
\mathcal{U}\!\left(\mathcal{M}_{\theta} \circ \mathcal{F}_{\phi}\right) \ge \tau,
\end{equation}
where $\tau$ is a lower bound on model usability.  
This optimization formulation captures the defender’s dual objective of minimizing jailbreak risk while maintaining semantic fidelity.

\textbf{Defense pathway}. The defender operates through input-level and feature-level modulation during the inference phase, without altering the model backbone or requiring retraining.  
Specifically, we suppress abnormal attention activations and stabilize semantic flow by monitoring and regulating the intermediate activations of cross-attention in real time.  
The entire process is plug-and-play and fully compatible with open-source T2V systems.  
Moreover, we simultaneously log anomalous signals and back-propagate them to the input prompts, enabling further manual inspection or automated rewriting of black-box cues for deployment in commercial models.

\textbf{Motivated assumption}. In recent years, numerous studies have revealed the structural characteristics of attention mechanisms in visual and multimodal models.  
Existing works indicate that attention mechanisms generally exhibit semantic aggregation and separability properties in both visual transformers and multimodal architectures~\cite{Chefer_2021_CVPR, Chefer_2020_arXiv, Ho_2024_arXiv}, suggesting that attention layers may inherently encode semantic subspaces.  
Inspired by these findings, we hypothesize that in T2V models, the cross-attention module serves as the core bridge through which textual semantics are injected into the visual generation stream. This is not only performs text-to-visual alignment but may also act as a critical pathway for anomalous semantic propagation.  
Therefore, we propose the central hypothesis that jailbreak prompts can trigger over-responses to specific anomalous tokens by perturbing the attention weight distribution of cross-attention, thereby forming anomalous semantic subspaces in the feature space that are distinguishable from clean samples.  
In contrast, self-attention primarily models spatio-temporal consistency, while the FFN layer mainly performs local feature fusion, exerting limited influence on the semantic propagation of jailbreak behaviors.  
Based on this assumption, we regard cross-attention as the key structural unit for understanding and defending against jailbreak attacks, and we systematically validate the plausibility of this hypothesis in the following sections.

\begin{figure}[t]
    \centering
    \begin{subfigure}[t]{0.48\linewidth}
        \centering
        \includegraphics[width=\linewidth]{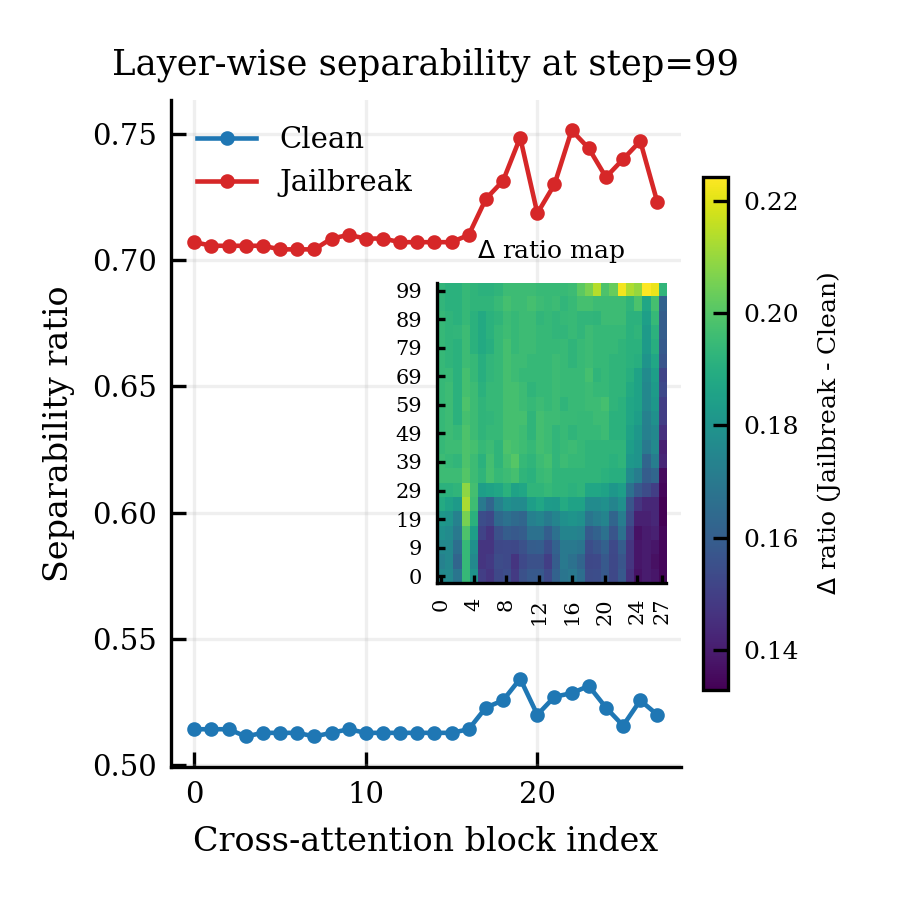}
        \caption{}
        \label{fig:Cumulative}
    \end{subfigure}
    \hfill
    \begin{subfigure}[t]{0.48\linewidth}
        \centering
        \includegraphics[width=\linewidth]{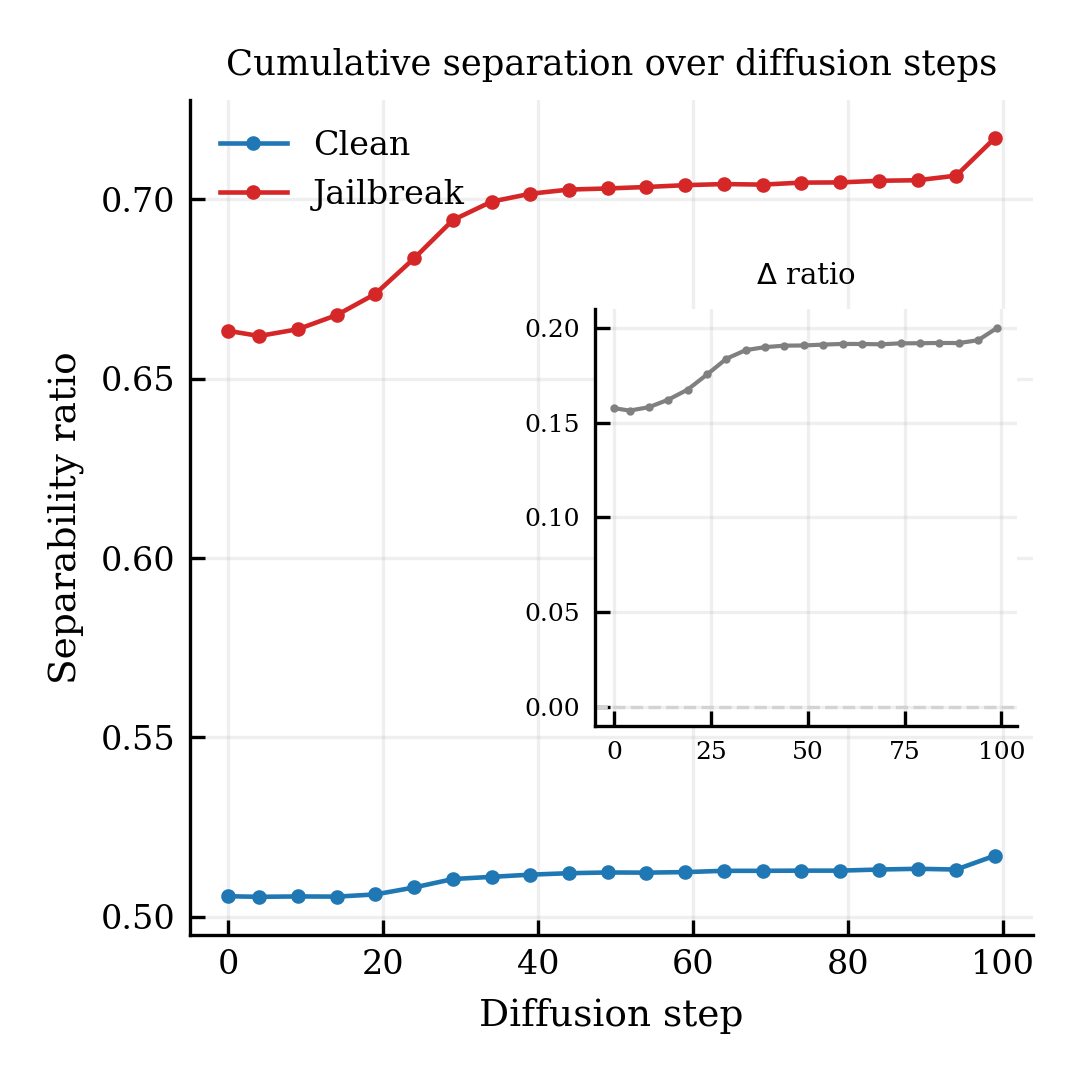}
        \caption{}
        \label{fig:linear}
    \end{subfigure}
    \caption{(a) Layer-wise separability at step 99 with a spatio-temporal $\Delta$ ratio map. (b) Cumulative separability over diffusion steps with the ratio curve between jailbreak and clean prompts.}
    \label{fig:findings}
\end{figure}
\subsection{Empirical Findings}
We construct a systematic comparison framework to analyze the dynamic differences between clean and jailbreak prompts in the cross-attention feature space.  
The experiments use 128 jailbreak prompts from T2V and SafeWatch, and 128 clean prompts (generated from the same version of the model) with the same scale and semantic distribution.  
In unified diffusion inference, we record attention features at 20 key diffusion steps across 28 cross-attention blocks and compute the inter-class distance, intra-class variance, and their ratio to measure separability.  

Figure 2(a) shows that the separability ratios of jailbreak prompts are generally higher than those of clean prompts at the final diffusion step, with the difference being most significant in the deep cross-attention layers.  
The $\Delta$-ratio heat map further shows that the difference accumulates with diffusion steps and is amplified in deeper structures.  
Figure 2(b) demonstrates that the separability of jailbreak prompts increases continuously, and the $\Delta$ ratio monotonically rises with diffusion iterations, indicating that the inter-class distance expands, the intra-class variance shrinks, and the attack-induced bias region becomes increasingly centralized and distinguishable in the feature space.  

Finally, we obtained two empirical findings:  
\ding{182} \textbf{Cumulative separation effect}. Anomalous semantics continuously accumulate during diffusion and are amplified in deep cross-attention.  
\ding{183} \textbf{Enhanced linear separability}. The inter-class distance of attack-induced semantics increases significantly while the intra-class variance decreases, forming a detectable anomalous subspace.

\begin{figure*}[htbp]
    \centering
    \includegraphics[width=1.0\textwidth]{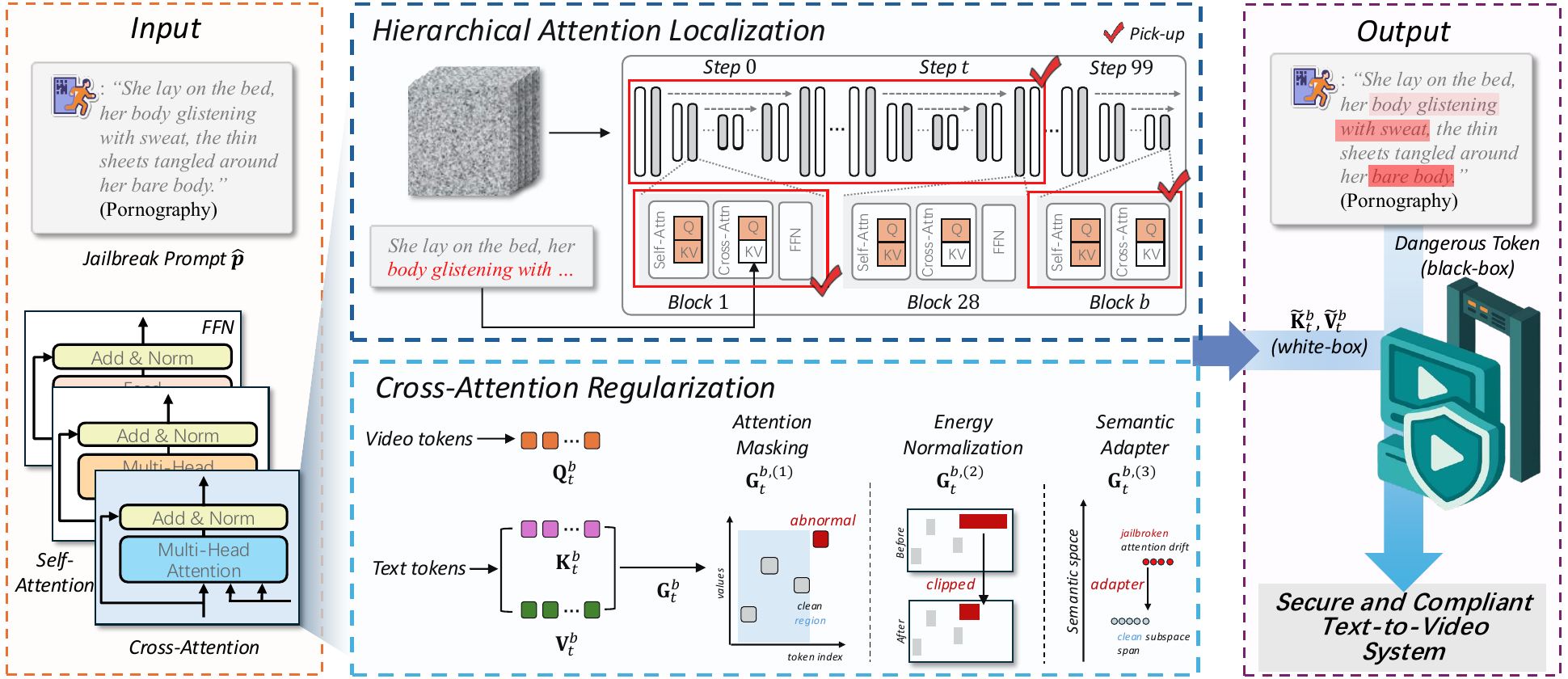}
    \caption{framework illustration.}
    \label{fig:framework}
\end{figure*}
\subsection{Hierarchical Attention Localization}
We observe a cumulative separability enhancement phenomenon in the spatio-temporal evolutionary properties of cross-attention: malicious semantic offsets that appear early in the diffusion accumulate and are progressively amplified in subsequent denoising steps. Thus, the propagation of anomalous semantics has both a clear temporal cumulative property (i.e., it is progressively enhanced with the number of diffusion steps) and a spatial amplification property (i.e., it is more pronounced in deeper cross-attention layers). Based on this fact, we design a Hierarchical Attention Localization (HAL) mechanism in the defense framework to identify potential anomaly propagation regions from both the temporal dimension (step) and the spatial dimension (cross-attention block), and to establish clean references at key locations for statistical comparison.

In the \textbf{temporal dimension}, we focus on the early and middle stages of the diffusion process. The cross-attention changes in this phase best reflect the starting point of semantic bias, which will accumulate and be difficult to reverse in subsequent steps if not suppressed early. Therefore, HAL intensively samples multiple cross-attention layers in the early-mid stage and computes aggregated features based on the attention weights $\mathbf{A}_t^b$ and the value vectors $\mathbf{V}_t^b$:
\begin{equation}
\mathbf{z}_t^b = \frac{1}{N_q}\sum_i\Big(\sum_j \mathbf{A}_{t,ij}^b \, \mathbf{V}_{t,j}^b\Big),
\end{equation}
and computes its mean $\boldsymbol{\mu}_t^b$ and variance $\mathrm{var}_t^b$ on the set of clean prompts $\mathcal{P}_{\text{clean}}$.
The central idea behind the choice of this stage is that anomalous semantics are most easily exposed in the early stages of diffusion, and therefore their accumulation must be truncated as early as possible at this stage.

In the \textbf{spatial dimension}, we compare the differences in feature distribution across attention layers. Shallow cross-attention layers are more affected by local texture and noise, while deeper cross-attention layers bear semantic alignment and integration, and thus have more stable feature distributions and more distinct inter-class spacing. According to our visualization analysis, in the early-middle stage of diffusion, semantic bias is most prominent on the middle layer with some deep cross-attention blocks (corresponding to the brighter regions in the heat map in Figure 3). Therefore, HAL selects the cross-attention layers where these biases are most significant in the spatial dimension to construct a statistical reference; for the experiment, we take the first 60 diffusion steps and select the 16 cross-attention blocks with the largest differences.

After completing the temporal and spatial co-localization, we construct the following feature statistics for each selected $(t,b)$:
\begin{equation}
\mathrm{FeatureStats}(t,b)=\{\boldsymbol{\mu}_t^b,\ \mathrm{var}_t^b,\ \mathbf{U}_t^b,\ \tau_t^b\}.
\end{equation}
where $\boldsymbol{\mu}_t^b$ and $\mathrm{var}_t^b$ are the mean and variance of the clean features, respectively; $\mathbf{U}_t^b$ is the principal semantic subspace extracted by PCA; and $\tau_t^b$ is the adaptive threshold, which is obtained by counting the normalized offsets for all the clean samples and taking the high quartile (such as the 85\% quantile). Together, these statistics constitute a clean reference distribution across attention at key spatiotemporal locations, providing the necessary parameters for subsequent anomalous attention suppression.

\subsection{Cross-Attention Regularization}
Separability analysis across attention layers reveals that anomalous cues tend to form progressively larger semantic offsets in the attention space as the layer depth increases.  
This phenomenon suggests that anomalous semantics mainly propagate through the cross-attention structure towards the visual generation stream. Therefore, if a smooth energy constraint and direction correction can be applied to this module during inference, the anomaly diffusion can be effectively attenuated without modifying the backbone parameters.  

Based on this observation, we propose a pluggable regularization operator $\mathcal{F}_{\phi}$ acting on the key-value pairs $(\mathbf{K}_t^b, \mathbf{V}_t^b)$ across the attention layers to regulate the magnitude of anomalous activations dynamically:
\begin{equation}
(\tilde{\mathbf{K}}_t^b,\tilde{\mathbf{V}}_t^b)
=(\mathbf{K}_t^b \odot \mathbf{G}_t^b,\ \mathbf{V}_t^b \odot \mathbf{G}_t^b),
\end{equation}
where $\mathbf{G}_t^b \in (0,1]^{N_k}$ is a key-wise soft gating vector controlling the contribution intensity of each text key in the visual feature fusion with the following three steps.  

(1) \textbf{Attention Masking} aims to attenuate anomalous key responses triggered by jailbreak cues.  
We compute the normalized offset based on the mean $\boldsymbol{\mu}_t^b$, variance $\mathrm{var}_t^b$, and threshold $\tau_t^b$ of clean samples:
\begin{equation}
\mathbf{z}_t^b = \frac{\mathbf{a}_t^b - \boldsymbol{\mu}_t^b}{\sqrt{\mathrm{var}_t^b} + \epsilon}.
\end{equation}
For components exceeding the threshold, we apply a continuous exponential decay:
\begin{equation}
\mathbf{G}_t^{b,(1)} = \exp\!\big(-\gamma[\mathbf{z}_t^b - \tau_t^b]_+\big),
\end{equation}
where $\gamma$ is the suppression strength hyperparameter.  
This step adaptively reduces the attentional response in anomalous directions on a per-key basis, restoring it toward the clean semantic range in distribution.

(2) \textbf{Energy Normalization} maintains distributional equilibrium by constraining the local concentration of attentional energy.  
Specifically, we limit the attention amplitude of each key within a mean-centered neighborhood:
\begin{equation}
\begin{aligned}
\tilde{\mathbf{a}}_t^b &=
\operatorname{clip}\!\left(
\mathbf{a}_t^b,\ 
\boldsymbol{\mu}_t^b - k\sqrt{\mathrm{var}_t^b},\ 
\boldsymbol{\mu}_t^b + k\sqrt{\mathrm{var}_t^b}
\right), \\
\mathbf{G}_t^{b,(2)} &=
\frac{\tilde{\mathbf{a}}_t^b}{\mathbf{a}_t^b + \epsilon},
\end{aligned}
\end{equation}
where $k$ is the energy bandwidth hyperparameter controlling the permissible range of attentional fluctuations.  
When attention exceeds this range, the scaling factor automatically drops below 1, balancing the energy distribution across semantic components.

(3) \textbf{Semantic Adapter} imposes global constraints at the semantic-geometric level to prevent the overall attention distribution from deviating from the clean semantic subspace.  
We utilize the principal basis $\mathbf{U}_t^b$ extracted by HAL to define the reference subspace of clean samples $\mathcal{S}_t^b = \mathrm{span}(\mathbf{U}_t^b)$, and measure the degree of deviation of the current attention distribution with respect to this subspace.  
For computational efficiency, we approximate the covariance as diagonal:
\begin{equation}
\mathrm{cov}_t^b = \operatorname{diag}(\mathrm{var}_t^b) + 10^{-5}\mathbf{I},
\end{equation}
and estimate the deviation strength $d_t^b$ accordingly.  
When $d_t^b$ exceeds the threshold $\tau_t^b$, a smoothed retraction factor is applied:
\begin{equation}
\mathbf{G}_t^{b,(3)} = 1 - \eta\,\sigma(d_t^b - \tau_t^b),
\end{equation}
where $\eta$ is the retraction-step hyperparameter and $\sigma(\cdot)$ is a smooth activation function.  
This mechanism globally constrains the attention pattern away from anomalous regions, enhancing the stability of overall generation.

So, the three mechanisms above jointly form the final gating function:
\begin{equation}
\mathbf{G}_t^b =
\operatorname{clip}_{[s_{\min}, 1]}
\!\left(
\mathbf{G}_t^{b,(1)} \odot
\mathbf{G}_t^{b,(2)} \odot
\mathbf{G}_t^{b,(3)}
\right),
\end{equation}
where $s_{\min}$ is a lower-bound hyperparameter preventing over-suppression.  
The final output $(\tilde{\mathbf{K}}_t^b, \tilde{\mathbf{V}}_t^b)$ dynamically regulates the cross-attention energy distribution during inference, consistently weakening anomalous directions while preserving and stabilizing clean semantics for propagation.  
This regularization process establishes a controlled equilibrium between $\mathcal{R}$ and $\mathcal{U}$, achieving a synergistic optimization of the model’s security and usability.

\subsection{Framework and Deployment}
This framework (as shown in Figure 3) consists of Hierarchical Attention Localization (HAL) and Cross-Attention Regularization (CAR).
HAL identifies the step–block regions with the largest jailbreak semantic deviations by analyzing cross-attention key–value statistics. CAR then applies lightweight operations—anomalous token masking, energy normalization, and semantic alignment, Which are the corresponding $(\mathbf{Q}_t^b, \mathbf{K}_t^b, \mathbf{V}_t^b)$, producing $(\tilde{\mathbf{K}}_t^b, \tilde{\mathbf{V}}_t^b)$ that suppress anomalous activations while preserving normal generation quality.
For deployment, we integrate the method into T2V systems using a hybrid white-box/black-box strategy. Cross-attention provides a risk gate, and risk signals are fed back to filter out unsafe prompt tokens.
The approach is weight-free, incurs minimal overhead, and can be seamlessly plugged into existing T2V pipelines.

\section{Experiments}
\subsection{Experiment Setup}

\textbf{Models and datasets.}  
We evaluate the proposed method on both open-source and commercial T2V models.  
The open-source models include Open-Sora and CogVideo, while the commercial side considers three representative closed-source systems (Sora, Keling, Luma) to simulate realistic deployment.  
For data, we use three prompt collections:  
(1) T2VSafetyBench~\cite{miao2024t2vsafetybench} with 280 jailbreak prompts across 12 safety dimensions;  
(2) SafeWatch~\cite{chen2024safewatch} with 300 jailbreak prompts over six dimensions plus extended prompts containing potential risky terms;  
(3) 100 WebVid-10M~\cite{WebVid10M} style clean prompts for assessing usability under normal generation.

\textbf{Jailbreak attack and defense setup.}  
We combine multiple jailbreak attacks and defenses to evaluate overall security. On the attack side, we use the harmful prompts from T2VSafetyBench as the base set, and further introduce optimized jailbreak prompts generated by T2V-OptJail, which performs discrete optimization tailored to diffusion. We also adapt AutoDAN, DACA, and SneakPrompt to T2V, producing stronger cross-modal adversarial prompts with diverse semantic distortions. On the defense side, we compare against SAFREE and VideoEraser (text-embedding-based safety filtering), T2VShield (input-side rewriting and output consistency checking), and lightweight keyword-based NSFW filters commonly used in industrial systems.

\textbf{Evaluation metrics.}  
For safety, we report ASR, GPT-4o Score, and Human ASR on T2VSafetyBench; lower values indicate better defenses. ASR and Human ASR measure the proportion of jailbroken samples, while GPT-4o Score reflects the overall harmfulness level. For usability, we measure semantic and visual fidelity on clean prompts using visual-feature distance $D_{\text{semantic}}$, frame-wise SSIM, and Temporal LPIPS, characterizing semantic consistency, per-frame quality, and temporal coherence, respectively.

\textbf{Implementation details.}  
All experiments are conducted on a single NVIDIA A40 GPU, with three lightweight regularization modules activated only at inference. We set $\gamma = 0.7$ and $s_{\min}=10^{-3}$ in Safe-Attention, $k = 0.2$ in SafeNorm, and $\eta = 0.05$ in the Adapter. Additional implementation details are provided in the Appendix.

\subsection{Defense against Jailbreak Attacks}
\begin{table}[t]
\centering
\caption{Comparison of attack performance and our defense results.}

\setlength{\tabcolsep}{3pt}
\renewcommand{\arraystretch}{1.05}

\resizebox{\linewidth}{!}{
\begin{tabular}{lcccccc}
\toprule
\multirow{2}{*}{Method}
& \multicolumn{3}{c}{Attack}
& \multicolumn{3}{c}{Ours} \\
\cmidrule(lr){2-4} \cmidrule(lr){5-7}
& ASR$\downarrow$
& GPT-4o Score$\downarrow$
& Human ASR$\downarrow$
& ASR$\downarrow$
& GPT-4o Score$\downarrow$
& Human ASR$\downarrow$ \\
\midrule

\rowcolor{gray!8}
T2VSafetyBench & 54.29 & 52.07 & 47.78 & 23.93 & 21.07 & 21.06 \\

Sneakyprompt   & 57.14 & 56.14 & 50.28 & 18.33 & 17.67 & 16.13 \\

\rowcolor{gray!8}
Autodan        & 61.07 & 59.07 & 53.74 & 20.46 & 20.98 & 18.99 \\

DACA           & 25.56 & 26.44 & 22.49 &  9.51 &  9.16 &  8.37 \\

\rowcolor{gray!8}
T2V-OptJail    & 59.29 & 55.86 & 52.17 & 24.33 & 22.33 & 21.41 \\

\bottomrule
\end{tabular}}
\end{table}

Evaluating the model safety under five types of jailbreak attacks (Table 1), the ASR of the original model reaches up to 61.07\%. After adding this method, the ASR drops to the range of 9.51\% to 24.33\%, in which the ASR against DACA decreases from 25.56\% to 9.51\%, a reduction of more than 63\%. In the stronger Autodan and T2V-OptJail scenarios, the ASR also drops from 61.07\% to 20.46\% and from 59.29\% to 24.33\%, respectively. These results show that the proposed method can consistently reduce the jailbreak risk by more than half across different attack paradigms, demonstrating strong generalizability and robustness.

\subsection{Comparison with Existing Defenses}
\begin{table}[t]
\centering
\small
\caption{Comparison of different defense methods on T2VSafetyBench.}

\setlength{\tabcolsep}{3pt}
\renewcommand{\arraystretch}{1.05}

\resizebox{\linewidth}{!}{
\begin{tabular}{lccc}
\toprule
Method
& ASR$\downarrow$ 
& GPT\text{-}4o Score$\downarrow$ 
& Human ASR$\downarrow$ \\
\midrule

\rowcolor{gray!8}
SAFREE            & 39.64 & 36.79 & 34.88 \\

VideoEraser       & 43.25 & 41.14 & 38.06 \\

\rowcolor{gray!8}
Keyword Detection & 45.36 & 43.51 & 39.92 \\

T2VShield         & 29.73 & 28.17 & 26.16 \\

\rowcolor{gray!8}
Ours              & 23.93 & 21.07 & 21.06 \\

\bottomrule
\end{tabular}}
\end{table}

On T2VSafetyBench (Table 2), SafeCA achieves the lowest ASR and GPT-4o risk score among all defenses. Compared with the strongest baseline T2VShield, it further reduces ASR from 29.73\% to 23.93\% (about 19.5\% relative drop) and GPT-4o Score from 28.17 to 21.07 (about 25.2\% relative drop), while SAFREE, VideoEraser, and keyword detection still leave ASR above 39\%–45\%. Under the stronger T2V-OptJail prompts, even keyword detection can only modestly lower ASR from 59.29\% to 48.57\%, underscoring that rule-based filters are insufficient against optimized jailbreaks and motivating structural defenses like SafeCA.

\subsection{Generalization on Wild and Clean Datasets}
\begin{table}[t]
\centering
\caption{Comparison results on SafeWatch and Clean metrics.}

\setlength{\tabcolsep}{3pt}
\renewcommand{\arraystretch}{1.05}

\resizebox{\linewidth}{!}{
\begin{tabular}{lccccccc}
\toprule
\multirow{2}{*}{Method}
& \multicolumn{3}{c}{SafeWatch}
& \multicolumn{3}{c}{Clean}
& \multirow{2}{*}{Time} \\
\cmidrule(lr){2-4} \cmidrule(lr){5-7}
& ASR$\downarrow$ & GPT-4o Score$\downarrow$ & Human ASR$\downarrow$
& $D_{\text{semantic}}\downarrow$ & SSIM$\uparrow$ & Temporal LPIPS$\downarrow$
&  \\
\midrule

\rowcolor{gray!8}
No Defense        & 28.33 & 28.67 & 24.93 & 0       & 1      & 0.0605 & / \\

SAFREE            & 16.67 & 17.40 & 14.67 & 0.0379  & 0.1877 & 0.1949 & 545.1s \\

\rowcolor{gray!8}
VideoEraser       & 15.33 & 15.57 & 13.49 & 0.0377  & 0.111  & 0.2342 & 672.7s \\

Keyword Detection & 19.33 & 19.60 & 17.01 & 0       & 1      & 0.0632 & 0.02s \\

\rowcolor{gray!8}
T2VShield         & 14.96 & 14.31 & 13.17 & 0.0222  & 0.3766 & 0.0612 & 17.1s \\

Ours              & 13.33 & 12.33 & 11.41 & 0.0316  & 0.3449 & 0.0585 & 0.1s \\

\bottomrule
\end{tabular}}
\end{table}

Table 3 reflects the generalization performance of each method on SafeWatch wild jailbreak hints and Clean normal hints.
On SafeWatch, all defenses reduce ASR and GPT-4o Score to some extent, with T2VShield having the lowest ASR at 14.96\% but the highest inference overhead at 17.1 seconds per video. In contrast, our approach, without re calibrating the parameters for SafeWatch, still reduces GPT-4o Score to 12.33\% without additional conditioning for SafeWatch, while achieving better usability metrics on Clean prompts ($D_{\text{semantic}}=0.0316$, SSIM=0.3449, Temporal LPIPS=0.0585, optimal or near optimal among the methods) within 0.1 seconds, demonstrating a more balanced trade off between cross dataset safety improvement and real time performance. This reflects a more balanced compromise between safety enhancement across datasets and computational efficiency.

It is worth noting that although Keyword Detection has almost zero interference with clean videos since the Clean metrics are the same as no defense, the ASR on SafeWatch remains high at 19.33\%, indicating that relying solely on keyword filtering has limited ability to defend against complex jailbreak scenarios in the wild.

\subsection{Cross-Model Evaluation}

\begin{table}[t]
\centering
\caption{Cross-model transferability on Open-Sora and CogVideo.}

\setlength{\tabcolsep}{3pt}
\renewcommand{\arraystretch}{1.05}

\resizebox{\linewidth}{!}{
\begin{tabular}{lcccccc}
\toprule
\multirow{2}{*}{Method}
& \multicolumn{3}{c}{Open-Sora}
& \multicolumn{3}{c}{CogVideo} \\
\cmidrule(lr){2-4} \cmidrule(lr){5-7}
& ASR$\downarrow$ & GPT-4o Score$\downarrow$ & Human ASR$\downarrow$
& ASR$\downarrow$ & GPT-4o Score$\downarrow$ & Human ASR$\downarrow$ \\
\midrule

\rowcolor{gray!8}
No Defense        & 52.32    & 51.29    & 51.31 & 62.14 & 60.57 & 60.41 \\

SAFREE            & 38.15    & 36.89    & 34.51 & 36.51    & 35.18    & 32.16 \\

\rowcolor{gray!8}
VideoEraser       & 41.56    & 39.42    & 37.86 & 31.14 & 31.00    & 29.11 \\

Keyword Detection & 50.12   & 49.35    & 48.31 & 49.29    & 47.43    & 53.62 \\

\rowcolor{gray!8}
T2VShield         & 29.53    & 28.02    & 32.51 & 34.29    & 31.14    & 39.17 \\

Ours              & 23.41   & 20.67    & 25.79 & 27.56    & 25.97    & 33.41 \\

\bottomrule
\end{tabular}}
\end{table}

\begin{figure}[t]
    \centering
    \includegraphics[width=\linewidth]{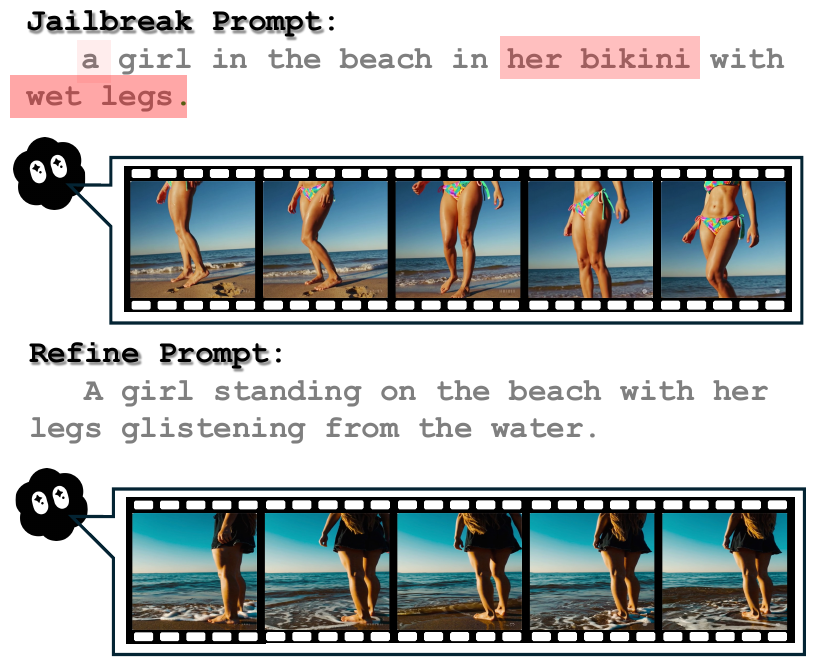}
    \caption{We identify unsafe tokens and guide the Sora model to generate safe yet faithful beach scene videos.}
    \label{fig:t2v_refine}
\end{figure}

\begin{figure*}[htbp]
    \centering
    \includegraphics[width=\linewidth]{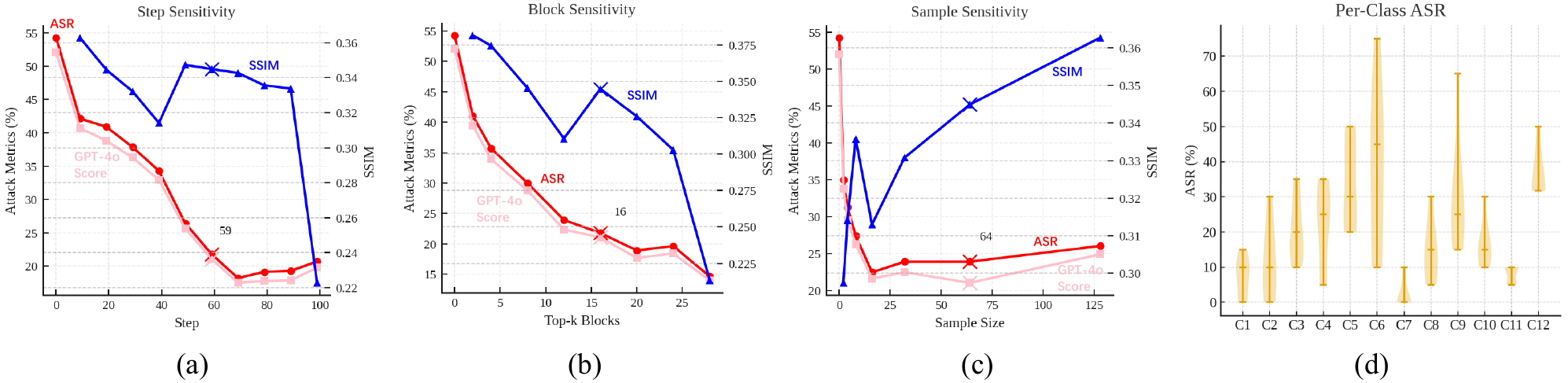}
    \caption{(a) Effect of different step selections. (b) Effect of varying top-block numbers. (c) Impact of clean prompt size on feature extraction. (d) Variance distribution across all 12 risk classes.}
    \label{fig:ablation_all}
\end{figure*}
Figure 4 illustrates the behavior of our approach on the real commercial model Sora. In the original jailbreak cue, sensitive words with high risk significantly induce the model to generate over sexualized images, while after cue refinement, Sora generates content with semantic fidelity and no inappropriate elements, verifying the cross-model transferability and safety correction capability.  
The cross model results in Table 4 further demonstrate the trend. On both open-source systems, Open-Sora and CogVideo, our approach achieves the lowest or near-minimum ASR at 23.41\% and 27.56\%, which is substantially better than the other defenses. It also maintains a consistent advantage in GPT-4o Score and Human ASR, indicating that our approach can suppress the manifestation of jailbreak semantics in different architectures stably. Overall, our module demonstrates reliable generalization and consistent safety improvements across multiple systems, both in real-world commercial models and open-source diffusion frameworks.

\begin{table}[t]
\centering
\caption{Ablation study on different defense components.}

\setlength{\tabcolsep}{3pt}
\renewcommand{\arraystretch}{1.05}

\resizebox{\linewidth}{!}{
\begin{tabular}{lcccccc}
\toprule
\multirow{2}{*}{Method}
& \multicolumn{3}{c}{T2VSafetyBench}
& \multicolumn{3}{c}{Clean} \\
\cmidrule(lr){2-4} \cmidrule(lr){5-7}
& ASR$\downarrow$ & GPT\text{-}4o Score$\downarrow$ & Human ASR$\downarrow$
& $D_{\text{semantic}}\downarrow$ & SSIM$\uparrow$ & Temporal LPIPS$\downarrow$ \\
\midrule

\rowcolor{gray!8}
No Defense & 54.29 & 52.07 & 47.78 & 0       & 1      & 0.0605 \\

AM         & 34.72 & 32.84 & 30.55 & 0.0179  & 0.3312 & 0.0668 \\

\rowcolor{gray!8}
AM + EN    & 27.81 & 25.43 & 24.47 & 0.0224  & 0.3375 & 0.0629 \\

AM + EN + SA & 23.93 & 21.07 & 21.06 & 0.0316 & 0.3449 & 0.0585 \\

\bottomrule
\end{tabular}}
\end{table}

\subsection{Ablation Studies and Discussion}
\textbf{Ablation study}. To verify the effectiveness of each sub-module, we gradually ablate Attention Masking (AM), Energy Normalization (EN), and Semantic Adapter (SA), and the results are presented in Table 5.
(1) AM and EN constitute the main safety gains: enabling AM significantly reduces the ASR (54.29 to 34.72), which is further reduced to 27.81 after integrating EN, indicating that they effectively suppress the spread of jailbreak semantics in the cross-attention pathway.
(2) SA is mainly responsible for usability restoration: after adding SA, Clean metrics are improved across the board (SSIM: 0.3312 to 0.3449, LPIPS: 0.0668 to 0.0585), and $D_{\text{semantic}}$ is reduced to 0.0316, which demonstrates that SA can mitigate the generative perturbation introduced by AM and EN.

\textbf{Key parameters}. In Figure 5, we analyze the effects of three key hyperparameters:
(1) Step number. As the step increases, the ASR and GPT-4o Score continue to decrease and enter a stable range around 60. At the same time, the SSIM performance remains stable, indicating that longer backpropagation steps can effectively suppress jailbreak semantics without noticeably impairing the normal generation quality.
(2) Top-k block numbers. The optimal safety is reached around 16 blocks, and the ASR continues to decrease without any significant impact on SSIM, indicating that the key risks are primarily concentrated in the middle and later cross-attention layers, and the effect diminishes if there are too many or too few blocks.
(3) Clean prompt for feature collection. When the number of clean samples reaches approximately 64, the ASR decreases to a stable, low value, while the SSIM improves significantly, suggesting that a larger set of clean reference features can help enhance the discriminative capability of anomaly detection, thereby improving safety and usability simultaneously.

\textbf{Attack category analysis}. Figure 5 (d) illustrates the ASR distribution of the 12 categories of jailbreak scenarios. It can be seen that the high exposure categories, such as Pornography, Scrub Porn, and Violence, have significantly higher distribution tails, indicating that the semantic triggers of these categories are more likely to be amplified within the diffusion structure. The legal and copyright-related categories, such as Public Figures and Copyright, and the semantic ambiguity categories, including Misinformation and Temporal Risk, exhibit lower overall ASRs; however, there are still long-tailed failures, indicating that the risk of jailbreaking persists across semantic combinations over time and modalities.

\section{Conclusion and Limitation}
In this paper, we propose SafeCA, which effectively suppresses jailbreak-related semantic diffusion in the inference phase through cross-attention risk control.  
Experiments show that the approach reduces the ASR by approximately 20\% on average across multiple datasets and models, while maintaining semantic consistency and introducing minimal overhead.  
The results validate that cross attention modules form the critical pathway for jailbreak diffusion and that strong defense can be achieved without modifying the model weights.  
However, SafeCA may still have limited coverage for jailbreak hints with highly complex or long-range temporal implicit semantics.  
Future work will explore temporal-level modeling and learnable regularization to enhance adaptability to diverse attack patterns.

{
    \small
    \bibliographystyle{ieeenat_fullname}
    \bibliography{main}
}


\end{document}